\documentclass{MITcsail}
\usepackage[
    backend=biber,
    style=numeric,
    sorting=none
  ]{biblatex}

\usepackage{hyperref}

\usepackage{cleveref}

\usepackage{adjustbox}
\usepackage{graphicx}

\title{Incorporating Emotions into Health Mention Classification Task on Social Media}

\author
{Olanrewaju Tahir Aduragba\footnote{Correspondence E-mail: olanrewaju.m.aduragba@durham.ac.uk}~$^{1, 2}$,  Jialin Yu~$^{1}$, Alexandra Cristea~$^{1}$\\
\vspace{1em} 
\normalfont{\small $^{1}$Department of Computer Science, Durham University, Durham DH1 3LE UK}\\
\normalfont{\small $^{2}$Department of Computer Science, Kwara State University, Kwara, Nigeria}\\
}

\begin{document}

\maketitle
\thispagestyle{empty}

\begin{abstract}
\textbf{ABSTRACT}\\
\emph{Introduction:}
The health mention classification (HMC) task is the process of identifying and classifying mentions of health-related concepts in text. This can be useful for identifying and tracking the spread of diseases through social media posts. However, this is a non-trivial task. Here we build on recent studies suggesting that using emotional information may improve upon this task. Our study results in a framework for health mention classification that incorporates affective features. 

\emph{Materials and Methods:} We present two methods, an intermediate task fine-tuning approach (\emph{implicit}) and a multi-feature fusion approach (\emph{explicit}) to incorporate emotions into our target task of HMC. We evaluated our approach on $5$ HMC-related datasets from different social media platforms including three from Twitter, one from Reddit and another from a combination of social media sources. The source code for our models is freely available at \url{https://github.com/tahirlanre/Emotion\_PHM}.

\emph{Results:} Extensive experiments demonstrate that our approach results in statistically significant performance gains on HMC tasks. By using the multi-feature fusion approach, we achieve at least a 3\% improvement in F1 score over BERT baselines across all datasets. We also show that considering only negative emotions does not significantly affect performance on the HMC task. Additionally, our results indicate that HMC models infused with emotional knowledge are an effective alternative, especially when other HMC datasets are unavailable for domain-specific fine-tuning.

\emph{Conclusion:} Our proposed framework incorporates emotional information into HMC tasks. The results validate that adding affective features improves the performance of detecting health information on social media, which is useful for epidemiological surveillance. As the HMC datasets come from various sources, including Twitter, Reddit and online health forums, this improvement demonstrates the generalisation and robustness of our framework. 
\end{abstract}


\section{Introduction}
Social media platforms such as Twitter and Reddit have become one of the most useful resources for people to share personal health experiences. This is partly due to their widespread availability and ease of accessibility. Also, the near real-time nature of these types of data makes them useful for public health surveillance. Despite this accessibility, there is a serious drawback to using these data sources. The large volume, the rate at which they are generated, and the unstructured nature of data pose special challenges. Potential biases may also exist with using such data \cite{filho2015twitter}. Overall, social media data is an important source of data and they have been shown to have applications in areas such as health informatics, public health, and medical research \cite{edo2020scoping}.


One of the crucial steps in harnessing social media data sources for public health surveillance is detecting content related to a health report. This task has been formulated as Health Mention Classification (HMC) \cite{biddle2020leveraging}. The HMC task aims to develop algorithms and models that can accurately identify and classify health mentions in a text (e.g. a social media post), enabling the automated analysis and interpretation of large volumes of health-related data. In this task, text documents are analyzed, and any mentions of health-related entities, such as diseases, symptoms, treatments, and other medical concepts, are identified and labelled according to a predefined set of categories or labels. HMC is a challenging problem due to the complex nature of natural language and the wide range of health-related entities that can be mentioned in a text.
For example, the post ``\emph{Every time I wrap gifts it looks good until I rapidly develop Parkinson's in both of my hands}" is a health mention, while ``\emph{Congratulations to Coach Parkinson on receiving a contract extension through the 2021-2022 season! \#JagsROAR}" is not. According to the former, the author suffers from Parkison's disease. In contrast, in the latter, the author refers to someone named Parkinson.

Previous work on HMC using social media has primarily focused on Twitter\footnote{https://twitter.com/} posts \cite{jiang2018identifying,karisani2018did}. Twitter is a popular data source for public health applications because its contents are mainly available to the public through the Twitter API. However, the 280-character limit on tweets can make it difficult to distinguish different contexts. More recently, Reddit posts have been collected for HMC tasks \cite{naseem2022identification}. Reddit posts are generally longer than tweets, allowing users to provide more context, and moderators often monitor discussions within subreddits to ensure relevance to the topic. This can make it easier to track conversations using health-related topics forums on Reddit. Reddit data, like Twitter, is publicly accessible through the Reddit API. Other dedicated online health forums such as \emph{AskaPatient}\footnote{https://www.askapatient.com/}, \emph{patient.info}\footnote{https://patient.info/} also exist for discussing health experiences. Such forums are used by patients to ask for medical advice from experts or to discuss with other patients.

To improve performance on HMC tasks, previous work has employed a variety of NLP techniques, from methods that use contextual word representations to non-contextual word representations \cite{biddle2020leveraging,karisani2018did}. Furthermore, past research has also considered modelling the literal or figurative usage of disease or symptoms words within texts expressing personal health experiences \cite{iyer2019figurative}. Another body of work also investigated using a combination of user behavioural information such as sentiment and emotion in conjunction with other features \cite{naseem2022identification}. However, in this work, we consider the relationship between self-reports of personal health experiences and emotions expressed in such reports. Health mentions are expected to trigger an emotion by the account poster. For example, someone reporting a diagnosis of a disease is likely to express emotions such as sadness, and fear, while someone who has recovered from an illness is likely to express emotions such as joy or happiness, even if they would be of similar health status, objectively; furthermore,  a post raising awareness about that particular disease might be neutral in terms of emotion. Our hypothesis is that the related nature of expressing emotions while discussing personal health experiences will result in a performance boost for our target task, i.e health mention classification in social media texts. 

In this paper, we explore the emotions conveyed in social media texts describing personal health experiences to improve our target task: health mention classification. To this end, we propose to \emph{implicitly} incorporate emotional knowledge into our target task through an intermediate emotion detection task.
We aim to leverage inductive bias from the emotion detection task to improve performance on HMC over baseline methods. We also propose modelling the relationship between emotions and health mentions by \emph{explicitly} combining HMC-specific and affective features to improve results further. We evaluate the effectiveness of our proposed approaches on five datasets from popular social media platforms such as Twitter and Facebook and online health communities. The datasets are of various sizes and characteristics. In addition, we investigate if there is any additional benefit to considering only negative emotions. 
Finally, we compare the performance of cross-task transfer from respective HMC models to a direct transfer from an emotion model.



\begin{table}[]
\centering
\caption{Distribution of labels and examples for each HMC dataset\label{tab:summary_datasets}}
\resizebox{.95\textwidth}{!}{%
\begin{tabular*}{\textwidth}{p{2cm}|p{2.5cm}p{1.5cm}p{10cm}}
\hline
\textbf{Dataset} & \textbf{Label} & \textbf{Size} & \textbf{Text} \\ \hline
\multirow{2}{*}{\textbf{FLU2013}} & Flu infection (positive) & 1,280 & \emph{$<$user$>$ Ugh. I'm getting a flu shot (hopefully) in about half an hour. :( Sorry yours is being ugly!}     \\ \cline{2-4} 
                         & Flu awareness (negative)  & 1,342 & \emph{I hope Is there some kind of flu going around? It's like everyone's getting sick all of a sudden. Weird. }    \\ \hline
\multirow{4}{*}{\textbf{PHM2017}} & Self-mention  & 306 &  \emph{Officially now a cancer patient (1991)}\\ \cline{2-4} 
                         & Other mention & 516 & \emph{Dana set a goal after her \#stroke: walk in high heels again $<$url$>$ \#2health \#ForOurHearts $<$url$>$ } \\ \cline{2-4} 
                         & Awareness &  1,278 &  \emph{\#Stroke threatens millions of lives. Learn the signs: $<$url$>$ \#ForOurHearts $<$url$>$}   \\ \cline{2-4} 
                         & Non-health mention &  2,483 &  \emph{You are Alzheimer's mascot you master of socialism $<$url$>$ }\\ \hline 
\multirow{3}{*}{\textbf{SELF2020}} & No self-disclosure  &  2,954  &  \emph{There is an otosclerosis community FB page which is quite helpful.} \\ \cline{2-4} 
                         & Possible self-disclosure &  2,586 &    \emph{Im basically taking one day at a time. I guess some viruses are unknown to medicine. So is what it is.} \\ \cline{2-4} 
                         & Clear self-disclosure & 1,010  &  \emph{Dementia and its Genetic Markers, many are known, however that may not mean you will end up with a problem. I have a Congenital Short Term Memory Defect from Birth, and I have had to relatives who died from Dementia.}  \\\hline       
\multirow{2}{*}{\textbf{ILL2021}}
& Negative &  18,435      &    \emph{Brain ‘pacemaker’ could prevent tremors and seizures for Parkinson’s and epilepsy sufferers }\\\cline{2-4} 
& positive  &    3,872        &  \emph{'I'm not OK': Michael Buble gets emotional talking about 5-year-old son's cancer battle}\\ \hline   
\multirow{3}{*}{\textbf{RHMD2022}} & Figurative mentions  &     3,430       &   \emph{Addiction to a Toy **As a kid, I was always addicted to this one toy called a slinky. I would spend hours and hours just fiddling with it. It seemed so satisfying to me. Whenever I would lose it, I would go into a depressing state for days and days, until I found it again. is it just me who has an addiction to a specific type of toy**.} \\ \cline{2-4} 
                         & Non-health mentions &    2,586        &    \emph{Court let Merck hide secrets about a popular drug’s risks - Lawsuits claim baldness drug Propecia causes sexual problems and depression. The judge sealed evidence suggesting the maker downplayed the side effects.} \\ \cline{2-4} 
                         & Health mentions &    3,360        &    \emph{I was diagnosed with Asperger's, OCD, Major depressive, and PTSD while I was inpatient. Ask me anything I was inpatient for 6 days due to homicidal thoughts and urges towards those who had hurt me emotionally and physically. And I put that hatred on others who did nothing wrong. In Inpatient I was diagnosed with Asperger's, OCD, and later after Outpatient, PTSD. I was abused by my mother, and three friends over the years. Physically and Mentally. Ask me anything. }\\\hline 
\end{tabular*}%
}
\end{table}

\section{Materials and Methods}\label{sec2}
\subsection{Data}
We explore a variety of HMC-related datasets from different social media platforms to study the general applicability of our approach. We use three Twitter datasets - \emph{FLU2013} \cite{lamb2013separating}, \emph{PHM2017} \cite{karisani2018did}, \emph{ILL2021} \cite{karisani2021contextual}, one Reddit dataset - \emph{RHMD2022} \cite{naseem2022identification} and one from a combination of Facebook, Reddit, Twitter and \emph{patient.info} - \emph{SELF2020} \cite{valizadeh2021identifying}. These datasets are annotated for classifying mentions of health-related concepts in social media text (e.g. \emph{health mention}/\emph{non-health mention} or \emph{flu infection}/\emph{Flu awareness}). Table \ref{tab:summary_datasets} presents the summary of all HMC datasets. More details about the construction and annotation are provided in \ref{appendixA}.


\subsection{Models}
\begin{figure}[]
\centering
\includegraphics[width=0.7\textwidth]{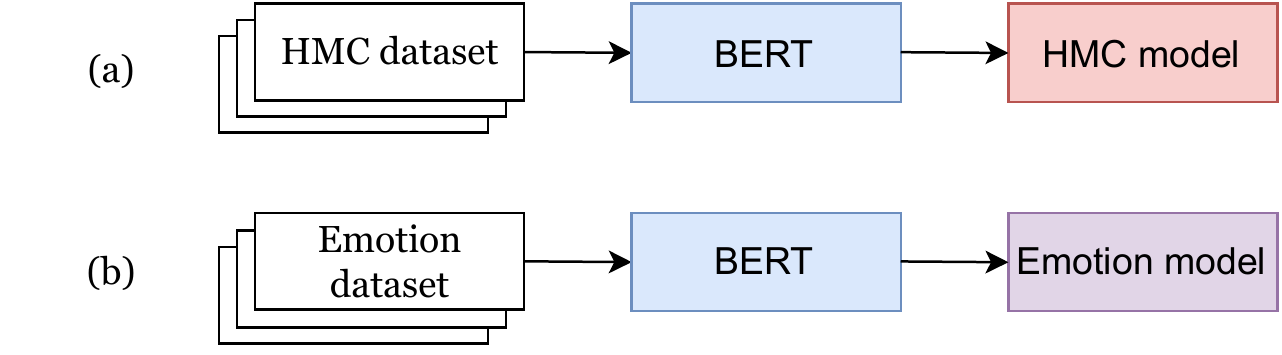}
\caption{The structure for (a) HMC model and (b) Emotion detection model }
\label{fig:model}
\end{figure}

Firstly, we describe our models for HMC and emotion detection (see Figure \ref{fig:model}). Following that, we present our framework for incorporating emotions into the task of HMC with two different types of enhancements. Both enhancements aim to enrich the neural representations learned by BERT with emotional knowledge.

\subsubsection{Health Mention Classification}
\label{sub:hmc_model}
Social media posts related to health mentions are usually extracted using symptoms or disease-related keywords. On the other hand, people on social media often use slang and varied representations of a word, which contribute to a high noisiness of social media posts. Thus, the presence of a symptom or disease word does not necessarily mean it is health-related. State-of-the-art approaches have been proposed to model contextual relationships between words in a text \cite{devlin2018bert}. Such models represent every word dependent on the particular context of occurrence. Incorporating contextual information is essential for language understanding tasks, even more so for correctly classifying health-related social media posts. 

To train our health mention classification model, we leverage a context-sensitive model, BERT \cite{devlin2018bert}. We fine-tune BERT on the respective HMC dataset. Fine-tuning BERT on task-specific corpus often yields performance gains on downstream tasks \cite{devlin2018bert}. We refer to the model fine-tuned on HMC dataset as $BERT_{HMC}$. 

\subsubsection{Emotion Detection}
\label{emotion_detection}
In computational linguistics, emotion detection involves identifying discrete emotions in text \cite{abdul2017emonet}. Many computational approaches have been proposed to detect emotions in text, including using emotion lexicons \cite{mohammad2012once} and machine learning models \cite{hasan2019automatic}. In line with the current trend in NLP, pre-trained language models have also been used to obtain state-of-the-art results in emotion analysis on social media \cite{desai2020detecting, aduragba2021detecting}. Moreover, it has been shown that understanding emotion requires a thorough understanding of context \cite{oatley2006understanding}. Due to this and the increasing ubiquity of pre-trained language models, we employ BERT \cite{devlin2018bert} for our emotion detection task. 

To capture fine-grained emotions, we leverage existing datasets annotated with emotions. We consider two publicly available emotion datasets - \emph{GoEmotions (GE)}
\cite{demszky2020goemotions} and \emph{SemEval18 - Emotions (SE) }\cite{SemEval2018Task1}. These datasets are manually annotated with various emotion categories such as \emph{anger}, \emph{disgust}, \emph{fear}, \emph{joy}, \emph{sadness} and \emph{surprise}. More information about the construction and labels of these datasets are provided in \ref{appendixB}. Based on standard practice in NLP, we fine-tuned BERT \cite{devlin2018bert} on an emotion dataset to learn general emotion representations. The domain-specific nature of emotion expressed in social media texts made this step crucial. We refer to the emotion model fine-tuned on \emph{GoEmotions} and \emph{SemEval18} as $BERT_{GE}$ and $BERT_{SE}$ respectively.  


\subsection{Emotion Incorporation Framework}
Studies have shown that social media users typically express a range of emotions when posting about personal health updates \cite{lerrigo2019emotional}. 
Building on this, we aim to capture the emotion spectrum when people post about their personal health experiences on social media. We consider two approaches to incorporate emotions into HMC. Both approaches aim to enrich the neural representations learned by BERT with emotional knowledge.

\subsubsection{Intermediate Task Fine-tuning Approach}
Recent work has shown that initially fine-tuning on an intermediate task before fine-tuning on a target task of interest improves the performance of pre-trained models \cite{phang2018sentence}. Notwithstanding, the effectiveness of this approach depends highly on the intermediate task that is applied \cite{chang-lu-2021-rethinking-intermediate}. The intuition behind intermediate fine-tuning is that if both tasks are related, the linguistic knowledge learnt in the intermediate task can contribute to understanding the target task. Following this observation, we hypothesize that emotion detection tasks can assist the task of HMC. 

\begin{figure} 
\centering
\includegraphics[width=0.6\textwidth]{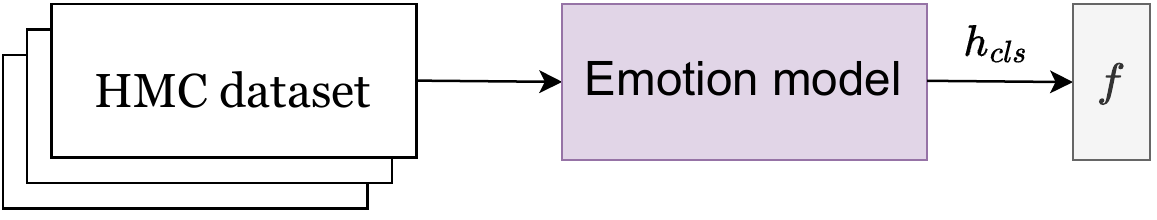}
\caption{\emph{Implicit} emotion incorporation with intermediate task fine-tuning }
\label{fig:intermediate_task}
\end{figure}

To improve upon our baselines, we explore intermediate fine-tuning as a means of \emph{implicitly} incorporating an emotion-specific inductive bias into our target task. We follow the emotion detection approach (\ref{emotion_detection}) described above to serve as an intermediate task. The intermediate fine-tuning step \emph{implicitly} learns affective features that could be helpful for the target task. Specifically, we use the fine-tuned emotion model parameters to initialise a new BERT model and then fine-tune the HMC task. The approach is illustrated in Figure \ref{fig:intermediate_task}.

\subsubsection{Multi-Feature Fusion Approach} \label{multi_feature}

\begin{figure}
\centering
\includegraphics[width=0.6\textwidth]{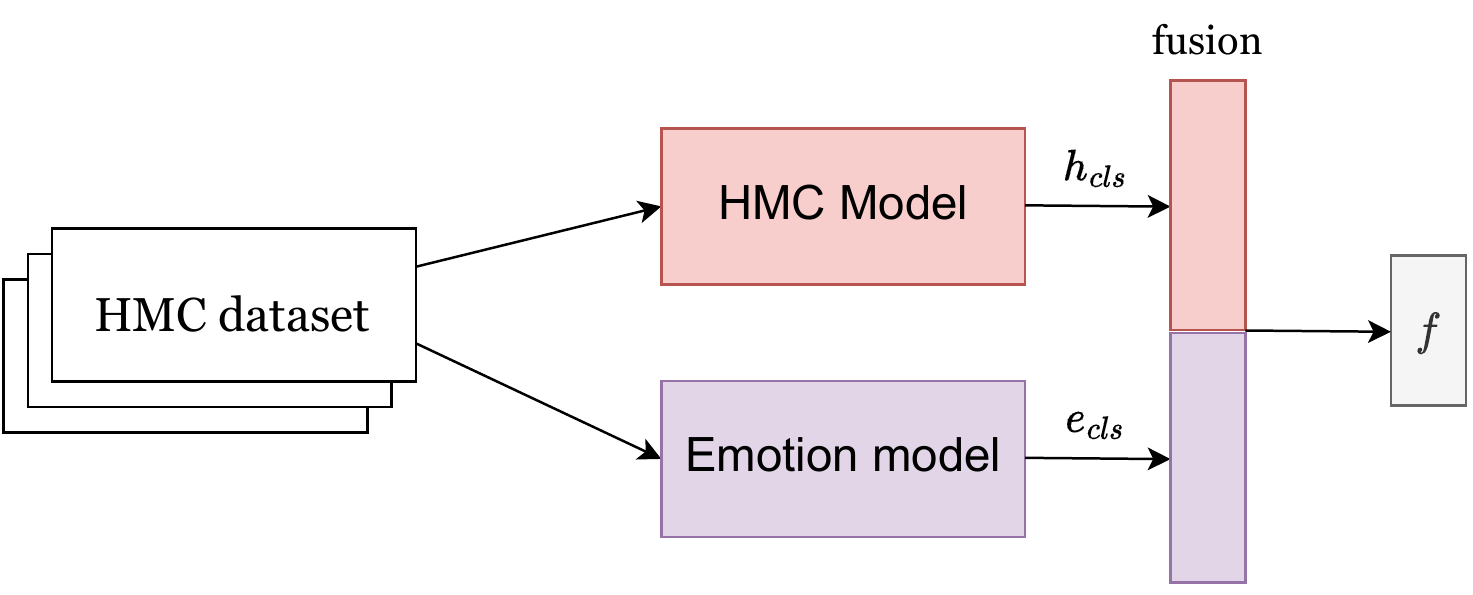}
\caption{\emph{Explicit} emotion incorporation with multi-feature fusion}
\label{fig:multi_feature}
\end{figure}

Research has demonstrated improved performance on HMC tasks when sentiment and emotional features are combined \cite{biddle2020leveraging, naseem2022identification}. They were generated, however, using emotion lexicons. Pre-trained language models capture better emotions expressed in social media texts due to their success in natural language understanding tasks \cite{sawhney2021phase}. We hypothesise that by combining emotional information, our health mentions specific sentence encoder could be guided to detect the nuances of reporting personal health experiences on social media. To achieve this, we \emph{explicitly} combine affective and HMC-specific linguistic features. We extract HMC-specific features from the HMC model and then these representations are fused with affective features extracted from the emotion model. In this approach, emotional information is incorporated \emph{explicitly} via the extracted affective features. The approach is illustrated in Figure \ref{fig:multi_feature}.

\subsection{Experimental Setup}\label{experiment}
\subsubsection{Model Architecture}

We describe our approach to fine-tune BERT for both HMC and emotion detection. Given an input sequence, we use a WordPiece tokenizer to tokenize the input as described in \cite{devlin2018bert}. The tokenizer adds two special tokens \emph{[CLS]} and \emph{[SEP]} to the input sequence, and the tokenized input is represented as: 

\begin{equation}
    X = [x_{[CLS]}, x_1, x_2, ...,x_n, x_{[SEP]}]
\end{equation}

where $x_t$ is the contextualised embedding of the $t$-th token in a sequence of $n$ symbols. The tokenized input is then passed into the BERT model to yield a sequence of hidden states as follows:

\begin{equation}
    H = [h_{[CLS]}, h_1, h_2, ...,h_n, h_{[SEP]}]
\end{equation}

We consider the hidden vector $h_{[CLS]} \in \mathbb{R}^{768}$ from the last hidden layer as the aggregate sequence representation for both HMC and emotion models. 
The representation is then passed through a linear output layer for prediction:
\begin{equation}
\tilde{y_i} = W_y h^i_{[CLS]} + b_y
\end{equation}

where $W_y$ and $b_y$ are learnable network parameters and $\tilde{y_i}$ is the network output. For the multi-feature approach (\ref{multi_feature}), the representations from both HMC and emotion models are directly concatenated\footnote{Initial experiments with max pooling and self-attention gave worse results.} to form a combined representation. Then, the fused representations are fed into a linear layer for prediction:
\begin{equation}
\tilde{y_i} = W_y (h^i_{[CLS]} \oplus e^i_{[CLS]}) + b_y
\end{equation}

where $h^i_{[CLS]}$ and $e^i_{[CLS]}$ are the extracted features from the HMC and emotion models, respectively. The HMC and emotion models are fine-tuned at the same time during training.

\subsubsection{Model optimization} 
For HMC tasks, which are single-label classification problems, the output uses a softmax activation and the network is optimized with a cross-entropy loss. For emotion tasks, which are multi-label classification problems, we use a sigmoid activation and optimize the network with a binary cross entropy loss.

\subsubsection{Baselines}
To serve as our baseline, we used the HMC model described above (\cref{sub:hmc_model}). Specifically, we use \emph{bert-base-uncased}, a version that consists of $12$ bidirectional transformers encoders with $768$ hidden layers and $12$ self-attention heads (total number of parameters = $110$M). We compare the performance of our emotion incorporation approach to the baseline.

\subsubsection{Training}
For all our experiments, we trained the models with minibatch gradient descent using the Adam optimizer \cite{kingma2014adam}. We used a batch size of $128$ (except for PHM2017\footnote{Initial experiments with batch size = 128 gave low performance.}, batch size = $64$). The number of epochs is set to 3 with a learning rate of $2e^{-5}$.

\paragraph{Dataset split} The dataset splits were not provided by the dataset distributors hence we create our own splits. For each dataset, we perform a \emph{80\%/10\%/10\%} split randomly to create the train, validation and test sets respectively. To train our models, the training set was used, while the validation set was used to select hyperparameters, and the test set was used to evaluate the performance of our models. The dataset splits we used for our experiments are presented in Table \ref{tab:data_split}.

\begin{table}[h]
\caption{Summary statistics of the dataset splits}\label{tab:data_split}
\centering
\begin{tabular}{lcccc}
\hline
\textbf{Dataset} & \textbf{Train} & \textbf{Validation} & \textbf{Test} & \textbf{All} \\ 
\hline
\textbf{FLU2013} & 2,098 & 263 & 264 & 2,622 \\ 
\textbf{PHM2017} & 3,667  & 459  & 460 & 4,583 \\
\textbf{SELF2020} & 5,241  & 656  & 656 & 6,550 \\ 
\textbf{ILL2021} & 17,846  & 2,232  & 2,232 & 22,307 \\
\textbf{RHMD2022} & 8,013  & 1,002  & 1,003 & 10,015 \\
\hline
\end{tabular}
\end{table}

\paragraph{Evaluation} Following previous works on HMC \cite{biddle2020leveraging,naseem2022identification}, we evaluate each model's performance using F1 macro score and report the results on the test set. To account for variability, we run each model five times with different seeds and report the average results over these five runs. We report the average performance across 5 runs with different seeds on the test set.

\section{Results}\label{results}
We show the results obtained using our framework in Table \ref{tab:all_results}. To determine whether the improvements are statistically significant, we use a two-sample t-test to compare the F1 scores. We assert significance if $p < 0.05$ under a two-sample t-test with the vanilla BERT model. Both of our approaches to incorporating emotional information boost performance across the HMC datasets. We also find that most of the gains are on the HMC datasets with limited samples.

For the intermediate task fine-tuning approach,  we observe that fine-tuning on either emotion dataset improves performance over the respective HMC task in the majority of cases. On some HMC datasets, such as FLU2013 and SELF2020, there was at least a 3\% increase in the performance. The BERT model fine-tuned on \emph{Sem-Eval18} emotion data ($BERT_{SE}$) yields the most improvements on all but one dataset, compared to the BERT model fine-tuned on GoEmotions ($BERT_{GE}$), which only obtained better improvement on PHM2017. Though, the vanilla BERT achieved better results on PHM2017 when compared to both $BERT_{GE}$ and $BERT_{SE}$. We note that $BERT_{SE}$ performs significantly better on RHMD2022 which is a Reddit-based dataset, than $BERT_{GE}$, which is trained on an emotion dataset from the same domain (Reddit). 

Overall, the results for the \emph{multi-feature} approach show the benefit of combining both health mention representations and emotional information. The \emph{multi-feature} approach consistently improves on the baseline and \emph{intermediate task fine-tuning} across all HMC datasets. The performance on the SELF2020 data shows the most significant improvement, up nearly 7 F1 points when emotion features are generated using $BERT_{SE}$ and up more than 7 F1 points when emotion features are generated with $BERT_{GE}$. Similarly to the results from our first approach, emotion-based models trained with \emph{Sem-Eval18} achieve the best performance in most cases.\\

\begin{table}[h]
\caption{F1 macro score for the health mention classification task. \textbf{Bold} denotes the highest score and * denotes statistical significance. The average of five random seeds is used for all scores.}\label{tab:all_results}
\centering
\begin{tabular}{lccccc}
\hline
\textbf{Model} & \textbf{FLU2013} & \textbf{PHM2017} & \textbf{SELF2020} & \textbf{ILL2021} & \textbf{RHMD2022}\\ 
\hline
\multicolumn{6}{c}{\emph{Baseline}} \\ \hline
\textbf{BERT$_{HMC}$} & 82.18 & 81.66 & 70.13 & 91.25 & 80.76 \\ \hline
\multicolumn{6}{c}{\emph{Intermediate Task Fine-tuning}} \\ \hline
\textbf{BERT$_{GE}$} & 82.18  & 81.29  & 73.02* & 91.32 & 80.91 \\
\textbf{BERT$_{SE}$} & 85.15  & 80.93  & 74.02* & 91.38 & 81.91* \\ \hline
\multicolumn{6}{c}{\emph{Multi-Feature Fusion}} \\ \hline
\textbf{BERT$_{HMC}$ + BERT$_{GE}$} & 85.85*  & 83.59*  & \textbf{77.28}* & 91.85* & 82.64* \\
\textbf{BERT$_{HMC}$ + BERT$_{SE}$} & \textbf{86.08}*  & \textbf{83.9}*  & 76.50* & \textbf{91.88}* & \textbf{82.77}* \\
\hline
\end{tabular}
\end{table}

\subsection{Effect of negative emotions}
Although a direct relationship has not been established, negative emotions have been associated with social media references to personal health \cite{wang2016twitter}. For example, tweets about colonoscopies were found to express more negative sentiment on average \cite{metwally2017using}. Another study showed that users post more frequently when symptoms are worse, raising concerns about bias towards negative emotions \cite{coulson2013online}. As a result, we investigate the effect of using only texts annotated with negative emotions to fine-tune our emotion model. We use a subset of the emotion dataset with only negative emotions (and neutral). For the GoEmotions data, we follow the negative emotions defined by the authors \cite{demszky2020goemotions} i.e. \emph{anger, disgust, fear, sadness, neutral}. For Sem-Eval18 - emotions data set, we used the following labels as negative emotions: \emph{anger, disgust, fear, pessimism, sadness}. Table \ref{tab:neg_emo_results} shows the result when we used only negative emotions.

\paragraph{Results} The results show no significant gain when using only negative emotions over using all emotions (positive, negative and neutral). This applies to both our approaches. The performance on most tasks deteriorated moderately for the \emph{intermediate task fine-tuning}. While there are improvements for our \emph{multi-feature} approach, these are relatively small and insignificant. This result shows no additional benefit to incorporating only negative emotions. Instead, taking advantage of the full spectrum of emotions might be more helpful.\\

\begin{table}[h]
\caption{F1 macro score for the health mention classification task. \textbf{Bold} denotes the highest score and * denotes statistical significance. The average of five random seeds is used for all scores.}\label{tab:neg_emo_results}
\centering
\begin{tabular}{lccccc}
\hline
\textbf{Model} & \textbf{FLU2013} & \textbf{PHM2017} & \textbf{SELF2020} &  \textbf{ILL2021} & \textbf{RHMD2022}\\ 
\hline
\multicolumn{6}{c}{\emph{Baseline}} \\ \hline
\textbf{BERT$_{HMC}$} & 82.18 & 81.66 & 70.13 & 91.25 & 80.76 \\\hline
\multicolumn{6}{c}{\emph{Intermediate Task Fine-tuning}} \\ \hline
\textbf{BERT$_{GE-neg}$} & 82.41  & 81.0  & 72.25* & 91.39 & 81.67* \\
\textbf{BERT$_{SE-neg}$} & 84.32  & 81.59  & 73.37* & 91.24 & 81.19 \\\hline
\multicolumn{6}{c}{\emph{Multi-Feature Fusion}} \\ \hline
\textbf{BERT$_{HMC}$ + BERT$_{GE-neg}$} & 86.07*  & 83.07*  & 76.94* & \textbf{91.93}* & \textbf{82.57}*\\
\textbf{BERT$_{HMC}$ + BERT$_{SE-neg}$} & \textbf{86.23}*  & \textbf{83.28}*  & \textbf{77.33}* & 91.91* & 82.42*\\
\hline
\end{tabular}
\end{table}

\subsection{Cross-HMC Task Transfer}
As part of our study, we compare the performance of using an emotion fine-tuned model to a model fine-tuned on a HMC data and transferred to other HMC tasks. For example, we fine-tune a bert-based model with PHM2017 and further fine-tune it on a target dataset, FLU2013. 

\paragraph{Results} We present the results obtained in Table \ref{tab:cross_task_results}. Here, we denote the best results between $BERT_{GE}$ and $BERT_{SE}$ as $BERT_{emotion}$. In some cases ($FLU2013$ and $PHM2017$), the model fine-tuned on emotion data, $BERT_{emotion}$ leads to better results than models fined-tuned on another HMC dataset. On the other datasets, the performance of the $BERT_{emotion}$ is very close to the best-performing fine-tuned models on HMC datasets. These findings demonstrate that we can use publicly available emotion datasets to enhance performance on HMC tasks without sparse annotated related HMC datasets.\\

\begin{table}[H]
\caption{F1 macro score for the health mention classification task. \textbf{Bold} denotes the highest score. The average of five random seeds is used for all scores.}\label{tab:cross_task_results}
\centering
\begin{tabular}{cccccc}
\hline
\textbf{Model} & \textbf{FLU2013} & \textbf{PHM2017} & \textbf{SELF2020} &  \textbf{ILL2021} & \textbf{RHMD2022}\\ 
\hline
\textbf{BERT$_{emotion}$} &  \textbf{85.15} & \textbf{81.29} & 74.02 & 91.38 &  81.91 \\
\textbf{BERT$_{FLU2013}$} & -   & 79.53  & 73.34  & 91.18  & \textbf{82.22}   \\
\textbf{BERT$_{PHM2017}$} &  84.98 & -  & \textbf{75.86} & \textbf{91.82} &  81.73  \\
\textbf{BERT$_{SELF2020}$} & 83.77  &  78.15 &  - & 91.57 & 81.57   \\
\textbf{BERT$_{ILL2021}$} & 84.52  & 77.99 & 74.16 & - & 82.0  \\
\textbf{BERT$_{RHMD2022}$} & 84.09 & 80.55 & 73.88 & 91.69 & - \\
\hline
\end{tabular}
\end{table}

\section{Conclusion}
In this paper, we showed that, as per our initial hypothesis, health mentions discussion contains emotional content, which can be exploited to improve health mention classification tasks. We proposed to incorporate emotions into HMC in two ways: (1) by implicitly adding affective features through intermediate fine-tuning on emotion detection task; and (2) by explicitly combining affective and HMC-specific features from both emotion and HMC models. Overall, we found that both approaches increased performance on the target task, with the explicit addition of affective features offering the highest gains (method 2). The benefits cut across all HMC datasets, demonstrating the generalisation and robustness of our approach. 

We also investigated if there is any relationship between negative emotions and health mentions. Our results show that there is no significant effect on the performance of HMC when only considering negative emotions for learning an emotion model. We further show that transferring emotion models to HMC tasks offers competitive performance to cross-HMC-task transfer. It suggests that in the absence of annotated data for HMC tasks, data-rich emotion tasks can be used to improve results.

Additionally, we observe that the HMC datasets with the least samples benefited most from the improvements. Given the fact that annotation size is a major bottleneck for health mention classification tasks, our methods can contribute to the wider community of researchers working in the HMC field. Our work points to numerous future directions, such as incorporating figurative language detection using our proposed approach and jointly modelling HMC and emotion detection tasks in a multi-task setting.

\section{Summary Points}
\emph{What was already known on the topic?}
\begin{itemize}
    \item Social media platforms have become important sources of information for public health surveillance.
    \item Health Mention Classification (HMC) is a task that involves identifying and classifying health-related mentions in text, to enable the automated analysis of large volumes of health data.
    \item Studies have explored the relationship between self-reported health experiences and emotional expression, and found that emotions can be useful features for HMC tasks.
\end{itemize}

\noindent \emph{What this study added to our knowledge?}
\begin{itemize}
    \item We propose two approaches to incorporate emotional information into HMC to improve performance.
    \item We achieved promising results and demonstrated the benefit of adding emotional information to detect health mentions from social media data, which can be useful public health surveillance.
    \item We demonstrated that data-rich emotion detection models can be transferred to HMC tasks with limited annotation data, resulting in competitive performance.
\end{itemize}




\printbibliography


\appendix
\section{Data}
\subsection{HMC Datasets}
\label{appendixA}
\paragraph{FLU2013} This dataset was created by \cite{lamb2013separating} to distinguish between reports of actual Flu infections and awareness. Each tweet in the dataset was manually annotated as either a \emph{flu report} (positive) or a \emph{flu awareness} (negative). At the time of this study, only 2,622 tweets were available to download, which is about 58\% of the original dataset.

\paragraph{PHM2017} Another existing dataset which focuses on more than one disease and condition was constructed by Karisani and Agichtein \cite{karisani2018did}. In the corpus, they collected English tweets related to Alzheimer’s disease, heart attack, Parkinson’s disease, cancer, depression, and stroke and manually annotated them in terms of \emph{self-mention}, \emph{other-mention}, \emph{awareness} and \emph{non-health}. At the time of this study, only 4,987 tweets were available to download, which is about 69\% of the original dataset. 

\paragraph{SELF2020} This dataset consists of health-related posts covering a range of health issues collected from online communities, including \emph{patient.info} and social media platforms (Facebook, Reddit and Twitter) \cite{valizadeh2021identifying}. The dataset consists of 6,550 posts annotated as either \emph{no self-disclosure}, \emph{possible self-disclosure} or  \emph{clear self-disclosure}. The majority (88.1\%) of the posts are from patient.info, thus the dataset contains phrases and sentences that are mostly longer than the Twitter-based datasets.

\paragraph{ILL2021} The ILL2021 dataset is an illness report dataset related to three different health conditions: Parkinson's disease, cancer, and diabetes \cite{karisani2021contextual}. The dataset is annotated to detect if a tweet mentions the health condition and contains a health report (\emph{positive}) or not (\emph{negative}). 22,307 tweets (98\% of the original dataset) were available for download at the time of this study.

\paragraph{RHMD2022} The RHMD dataset focuses on Reddit posts only \cite{naseem2022identification}. The posts contain keywords related to up to 15 diseases and symptoms such as Headache, OCD and Allergic. In total, the dataset contains 10,015 unique posts. 
They are labelled with \emph{figurative mention, non-personal health mention and health mention}. 
In terms of length, the posts are longer than the Twitter-based datasets. 

\subsection{Emotion Datasets}
\label{appendixB}

\paragraph{GoEmotions (GE)}  GoEmotions \cite{demszky2020goemotions} is a benchmark emotion dataset originally annotated with 27 diverse emotions and neutral. The dataset contains 58,000 Reddit comments. The authors further group the labels into 6 Ekman emotion groups and neutral. This is the variant we use for our experiments,  where emotions = \{\emph{anger, disgust, fear, joy, sadness, surprise and neutral}\}.

\paragraph{SemEval18 - Emotions (SE)} This dataset, which includes emotion-specific labels indicating the authors' emotional states, was taken from SemEval-2018 Task-1 \cite{SemEval2018Task1}. It comprises 10,986 tweets divided into 11 emotion labels – (\emph{anger, disgust, anticipation, fear, joy, love, optimism, pessimism, sadness, surprise and trust}), each of which is a binary label that denotes the presence of a specific emotion.

\beginsupplement

\section{Supplementary Materials}
Information about related work are provided in a supplementary file attached to this submission.

\end{document}